\documentclass{article}
\usepackage{ijcai20}

\usepackage{times}
\usepackage{soul}
\usepackage{url}
\usepackage[hidelinks]{hyperref}
\usepackage[utf8]{inputenc}
\usepackage[small]{caption}
\usepackage{graphicx}
\usepackage{amsmath}
\usepackage{amsthm}
\usepackage{booktabs}
\usepackage{algorithm}
\usepackage{algorithmic}
\usepackage{tabu}
\usepackage{multirow}
\usepackage[table,xcdraw]{xcolor}
\usepackage{cleveref}
\crefname{section}{§}{§§}
\Crefname{section}{§}{§§}

\title{IITK at the FinSim Task: Hypernym Detection in Financial Domain via Context-Free and Contextualized Word Embeddings}

\author{
Vishal Keswani$^{*}$
\and
Sakshi Singh\thanks{\quad Authors equally contributed  to this work.}  \qquad
\And
Ashutosh Modi
\affiliations
Indian Institute of Technology Kanpur (IITK)\\
\emails
\{vkeswani,sakshia\}@iitk.ac.in,
ashutoshm@cse.iitk.ac.in
}

\begin{document}

\maketitle

\begin{abstract}
In this paper, we present our approaches for the FinSim 2020 shared task on ``Learning Semantic Representations for the Financial Domain". The goal of this task is to classify financial terms into the most relevant hypernym (or top-level) concept in an external ontology. We leverage both context-dependent and context-independent word embeddings in our analysis. Our systems deploy Word2vec embeddings trained from scratch on the corpus (Financial Prospectus in English) along with pre-trained BERT embeddings. We divide the test dataset into two subsets based on a domain rule. For one subset, we use unsupervised distance measures to classify the term. For the second subset, we use simple supervised classifiers like Naive Bayes, on top of the embeddings, to arrive at a final prediction. Finally, we combine both the results. Our system ranks $1^{st}$ based on both the metrics, i.e., mean rank and accuracy. 
\end{abstract}

\section{Introduction} 
Natural Language Processing has mainstream applications in a wide range of domains. In the Financial domain, sentiment analysis is vastly simplified, while applications like financial document processing remain relatively unexplored. According to the popular educational website Investopedia\footnote{https://www.investopedia.com/terms/p/prospectus.asp}, ``A prospectus is a formal document that is required by and filed with the Securities and Exchange Commission (SEC) that provides details about an investment offering to the public." The ease in availability of financial texts in the form of Financial Prospectus opens a broad area of domain-specific research for computational linguists and machine learning researchers.   

A hypernym is simply a word (or concept) denoting a superordinate category to which words (or concepts) having more specific meaning belong. Hypernym detection is a relatively old problem studied in NLP for more than two decades \cite{shwartz2016improving}. It finds applications in question answering \cite{yahya2013robust}, web retrieval, website navigation or records management \cite{Bordea2015SemEval2015T1} and taxonomy evaluation \cite{yu2015learning}. In cognitive science, hypernyms are analogous to higher levels of abstraction in the hierarchy within which we innately organize concepts. Any concept at a lower level can be categorized as a hyponym while the corresponding higher-level concept is its hypernym. A hyponym can be associated with multiple hypernyms (Labrador: Dog, Animal; Revenue Bond: Bond, Security). Hence, hyponym-hypernym pairs are associated with a kind of `is-a' relationship.        

The problem of discovering suitable hypernyms has been formulated in different ways in the past. Previously, the SemEval community has organized similar tasks under the umbrella of taxonomy evaluation \cite{Bordea2015SemEval2015T1,bordea2016semeval}. The problem can also be proposed as a binary verification task, i.e., given a pair of terms, find whether they form a hypernym-hyponym pair. Most recently in SemEval-2018, the problem was reformulated as given a hyponym, find candidate hypernyms in a domain-specific search space \cite{camacho2017babeldomains}. The FinSim task \cite{finsim-task:2020} is perhaps the first hypernym detection task in the Financial domain. The problem is devised as a multi-class classification task. Each financial term (hyponym) is classified into one of the eight high-level classes (hypernym), which are mutually exclusive from each other.

In section \ref{sec:relatedwork}, we provide a brief literature review of the work already done in this field. In section \ref{sec:methods}, we describe the techniques used in our systems including word-embeddings and classifiers. In section \ref{sec:expsetup}, we discuss the experimental setup. This includes the systems that we submitted along with post-submission analysis. Section \ref{sec:results} summarises the results of all the systems. Finally, we conclude the paper in section \ref{sec:conclusion} and suggest future directions for research.

\section{Related Work}
\label{sec:relatedwork}
The literature on modelling hypernymy can be classified into two broad categories: Pattern-based and Distributional. Pattern-based approaches rely on the co-occurrence of hyponym and hypernym \cite{grefenstette2015inriasac}, substring matching, lexico-syntactic patterns \cite{lefever2014hypoterm} or organizing terms in a hierarchy or directed acyclic graph \cite{velardi2013ontolearn}.

Distributional approaches are relatively recent. Distributional approaches capture far away relationships and, unlike the pattern based approaches, do not rely on the co-occurrence of hyponym and hypernym in text. A typical model uses a distributional representation of a word also called word-embedding, as input for a classification layer \cite{santus2014chasing,fu2014learning,weeds2014learning,espinosa2016supervised,nguyen2017hierarchical}. 

Shwartz et al. \shortcite{shwartz2016improving} combined both pattern-based and distributional approaches in a neural network based model. Bernier-Colborne and Barriere \shortcite{bernier2018crim} use a combination of embeddings and Hearst-style patterns for hypernym detection. We leverage both the approaches in our analysis. We test for string inclusion to divide the dataset into two subsets. We then perform separate analysis on the subset of terms that include a class label and the subset of terms that exclude any class label or include multiple labels.

\section{Methods}
\label{sec:methods}
We employed a variety of methods that were essentially distributional. Figure \ref{fig:flowchart} shows a typical system. It primarily consists of an embedding layer followed by a classification layer. We discuss both the layers below.  

\subsection{Word-embeddings}
We employ two types of word-embeddings. One is based on the context-free Word2vec model \cite{mikolov2013efficient}. The second is the contextualized state-of-the-art language representation model, BERT \cite{devlin2018bert}.  
\subsubsection{Context-free embeddings: Word2vec}
We use Word2vec embeddings \cite{mikolov2013efficient} for capturing semantic and syntactic properties of words. It is a dense low-dimensional representation of a word. We trained the embeddings on the whole corpus of Financial Prospectus. Word2Vec represents each word as a vector. We tried different dimensions ranging from 50 to 500. A term is represented by an average of word embeddings of each word contained in the term. 
\subsubsection{Contextualized embeddings: BERT}
Bidirectional Encoder Representations from Transformers (BERT) \cite{devlin2018bert} is the state-of-the-art language model, that has been found to be useful for numerous NLP tasks. It is deeply bidirectional (takes contextual information from both sides of the token) and learns a representation of text via self-supervised learning. BERT models pre-trained on large text corpora are available, and these can be trained for a specific NLP task or further fine-tuned on a specific corpus. We used BERT Base Uncased configuration\footnote{https://github.com/google-research/bert}, which has 12 layers (transformer blocks), 12 attention heads and 110 million parameters. We extract sentences from the corpus containing the terms in train and test datasets (maximum 5 for each term). We extracted the default pre-trained embeddings from the last hidden layer for each word in a sentence. We obtain term-embeddings by taking an average of embeddings of its constituent words. This way, we get multiple embeddings for the same term. They are again combined by taking an average. We have limited access to computational resources, however, with higher computational capability, BERT can be fine-tuned on the whole corpus before extracting embeddings.  

\begin{figure}[]
\centering
\includegraphics[trim= 210 130 0 70, clip, scale=0.65]{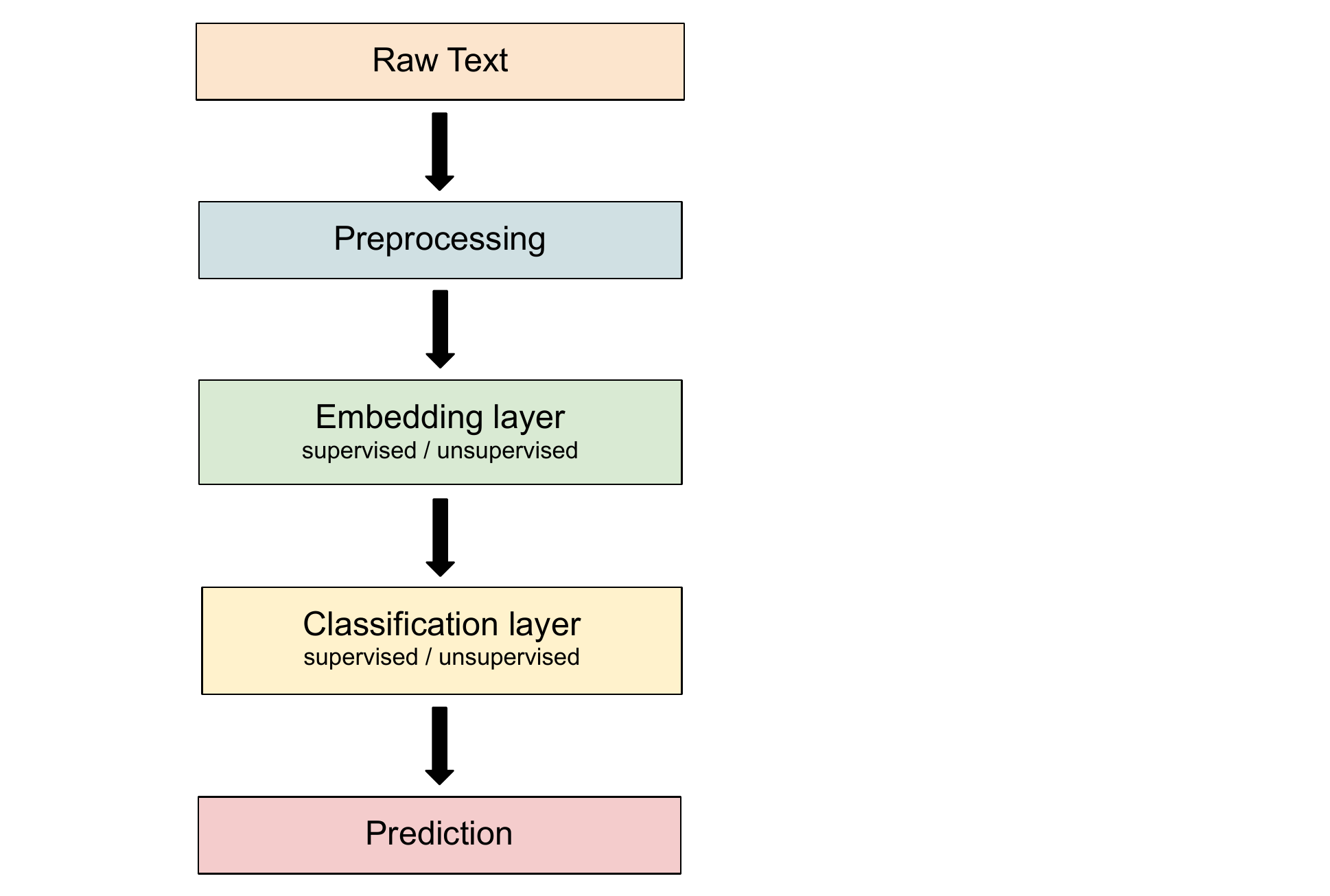}
\caption{A typical system pipeline}
\label{fig:flowchart}
\end{figure}

\subsection{Classifiers}
We use simple classifiers on top of the embedding layer. This is due to the small size of the train and test datasets (roughly 100 terms each). We perform both supervised and unsupervised classification. 

\begin{table*}[]
\centering
\renewcommand{\arraystretch}{1.35}%
\begin{tabular}{|c|c|c|c|c|c|}
\hline
\multicolumn{2}{|c|}{\textbf{Train}}      & \multicolumn{2}{c|}{\textbf{Test}}        & \multicolumn{2}{c|}{\textbf{Example}} \\ \hline
\textbf{\# hypernyms within term} & \textbf{\# terms} & \textbf{\# hypernyms within term} & \textbf{\# terms} & \textbf{Term}   & \textbf{Hypernym}   \\ \hline
0 & 41  & 0 & 29 & Debenture          & Bonds  \\ \hline
1 & 53  & 1 & 66 & Covered Bond          & Bonds  \\ \hline
2 & 6   & 2 & 4  & Bond Future & Future \\ \hline
  & 100 &   & 99 &                          &        \\ \hline
\end{tabular}
\caption{Distribution of terms as per hypernym inclusion}
\label{tab:dataset table}
\end{table*}

\subsubsection{Unsupervised classification} We obtain embeddings both for the terms and the 8 class labels.  We test three different measures of distance/similarity. First, we use cosine-similarity. It is a measure of similarity between two vectors. The more the cosine-similarity score, the closer are the embeddings of the term and the label. We rank the labels in descending order of similarity. We then employ two distance measures, L1 and L2, to find the distance between embeddings of the term and the class labels. The smaller the distance, the closer is the term to that class label. We rank the labels in the ascending order of distance for prediction. These measures do not depend on the size of the dataset as they do not involve further training in the classification layer.     
\subsubsection{Supervised classification}
We test two simple supervised classifiers, namely Na\"ive Bayes and Logistic Regression. Na\"ive Bayes is a popular classical machine learning classifiers \cite{rish2001empirical}. The primary assumption behind the model is that given the class labels. All features are conditionally independent of each other, hence the name Na\"ive Bayes. It is highly scalable, that is, takes less training time. It also works well on small datasets. We used the default Bernoulli Na\"ive Bayes classifier from sklearn library\footnote{https://scikit-learn.org/stable/}. Since the embeddings used are continuous, we first tried Gaussian Naive Bayes. However, the results were unsatisfactory. We then tried Bernoulli Naive Bayes. It binarizes the continuous embeddings with a default threshold of 0. It performed best with the default threshold, far better than Gaussian Na\"ive Bayes. In the following paper, we address Bernoulli Na\"ive Bayes as simply Na\"ive Bayes. 

Logistic regression \cite{kleinbaum2002logistic} uses logistic function as the representation, in a manner similar to linear regression, to model a binary dependent variable. The eight classes are treated as eight binary variables, which are assigned a probability between 0 and 1. Being a simple model, it works pretty well on small datasets. We use the default Logistic Regression classifier from the sklearn library. 

\section{Experimental Setup}
\label{sec:expsetup}
In this section, we quantitatively describe the dataset provided by the organizers and the challenges accompanying it. We then mention the preprocessing steps briefly. Finally, we discuss the architecture and parameters of the systems in detail. 


\subsection{Data description}
\label{subsec:datadescp}
As a part of the task, we are provided with a training dataset of 100 terms with corresponding class labels (hypernyms). The test dataset comprised of 99 financial terms. As a common observation, the majority of the terms contained the label within them (Table \ref{tab:dataset table}). For instance, consider the term ``Convertible Bonds". The corresponding label for this term is ``Bonds". Hence, such terms can be separately dealt using a rule-based approach. The text corpus provided by the organizers consisted of 156 Financial Prospectuses in PDF. 

The dataset (Table \ref{tab:dataset table}) comes with a lot of inherent challenges. Firstly, the dataset is too small for a supervised approach, especially neural network classifiers. Secondly, there were some terms in the training data, which were not present in the provided corpus. Also, the corpus was provided as PDFs and converting them to txt format added much noise and sentence boundary detection proved to be a challenge.

Another issue is related to acronyms. In both train and test datasets, there were multiple terms written as acronyms. For example, the term ``CDS" stands for Credit Default Swap. If the full form was given, this term would have easily qualified for subset 1, and direct classification would have been possible. However, because of the acronym form, the correct classification is solely dependent on the presence of ``CDS" in the corpus. The constituent terms Credit, Default and Swap also cannot be used to classify it.

\subsection{Data preprocessing}
Text preprocessing steps included removal of punctuation, stop words and special characters, followed by lower-casing, lemmatization and tokenization. We used the nltk library\footnote{https://pythonspot.com/category/nltk/} \cite{loper2002nltk} for the same. The tokens were then converted to vectors using Word2vec or BERT embeddings. Finally, the average of all the word vectors is taken to create final embedding for each term.

\subsection{Systems} 
\label{subsec:systems}
 As mentioned in section \ref{subsec:datadescp}, some of the terms contained the label within them. We split the test dataset into two subsets. Subset 1 consists of terms containing exactly one class label within them. Subset 2 has the remaining terms, those with no class label or more than one class label. Subset 1 and subset 2 comprise of 66 and 33 terms, respectively. We perform a separate analysis on both subsets.  On observing the training dataset, the terms in subset 1 can be directly classified into the corresponding label since they contain the label within them. This rule-based approach, of directly classifying a term into the label, works very well for our dataset with 100\% accuracy. But it does not provide a ranking of labels useful in potential exceptional cases for which the label contained in the term might not be the correct label. Though no such example is encountered in our dataset of 199 points in total, we do not have evidence to eliminate the possibility in which the ranking would be useful in evaluation according to mean rank. Hence, we run all the approaches used for subset 2 on subset 1 also. 



A typical system is represented in Figure \ref{fig:flowchart}. The combination of the classification layer and the embedding layer used in subset 1 and subset 2 may vary for each system. We describe five such combinations for subset 1 and subset 2. Both are combined to obtain results on the complete test dataset. The results for these systems are discussed in section \ref{sec:results}.

\begin{table*}[]
\centering
\renewcommand{\arraystretch}{1.35}%
\begin{tabular}{|c|c|c|c|c|c|c|c|c|c|c|c|c|c|}
\hline
\multicolumn{2}{|c|}{\textbf{}} &
  \multicolumn{6}{c|}{\textbf{Unsupervised}} &
  \multicolumn{6}{c|}{\textbf{Supervised}} \\ \hline
\multicolumn{1}{|c|}{\multirow{2}{*}{\textbf{Embedding}}} &
  \multicolumn{1}{c|}{\multirow{2}{*}{\textbf{Dim}}} &
  \multicolumn{2}{c|}{\textbf{Cosine Sim.}} &
  \multicolumn{2}{c|}{\textbf{L1}} &
  \multicolumn{2}{c|}{\textbf{L2}} &
  \multicolumn{3}{c|}{\textbf{Na\"ive Bayes}} &
  \multicolumn{3}{c|}{\textbf{Logistic Regression}} \\ \cline{3-14} 
\multicolumn{1}{|c|}{} &
  \multicolumn{1}{c|}{} &
  \textbf{MR} &
  \textbf{ACC} &
  \textbf{MR} &
  \textbf{ACC} &
  \textbf{MR} &
  \textbf{ACC} &
  \textbf{\#train} &
  \textbf{MR} &
  \textbf{ACC} &
  \textbf{\#train} &
  \textbf{MR} &
  \textbf{ACC} \\ \hline
\multirow{3}{*}{Word2vec} & 50  & 1.06 & 0.95 & 1.04 & 0.95 & 1.06 & 0.95 & 100 & 1.47 & 0.73 & 100 & 1.26 & 0.85 \\ \cline{2-14} 
                          & 100 & 1.02 & 0.98 & 1.02 & 0.98 & 1.00 & 1.00 & 100 & 1.21 & 0.85 & 100 & 1.11 & 0.89 \\ \cline{2-14} 
                          & 300 & 1.00 & 1.00 & 1.00 & 1.00 & 1.00 & 1.00 & 100 & 1.04 & 0.97 & 100 & 1.03 & 0.97 \\ \hline
BERT                      & 768 & 1.21 & 0.95 & 1.21 & 0.95 & 1.21 & 0.95 & 100 & 1.04 & 0.98 & 100 & 1.00 & 1.00 \\ \hline
\end{tabular}
\caption{Performance on subset 1 (MR = mean rank, ACC = accuracy)}
\label{tab:subset1}
\end{table*}

\begin{table*}[]
\centering
\renewcommand{\arraystretch}{1.35}%
\begin{tabular}{|c|c|c|c|c|c|c|c|c|c|c|c|c|c|}
\hline
\multicolumn{2}{|c|}{\textbf{}} &
  \multicolumn{6}{c|}{\textbf{Unsupervised}} &
  \multicolumn{6}{c|}{\textbf{Supervised}} \\ \hline
\textbf{Embedding} &
  \textbf{Dim} &
  \multicolumn{2}{c|}{\textbf{Cosine}} &
  \multicolumn{2}{c|}{\textbf{L1}} &
  \multicolumn{2}{c|}{\textbf{L2}} &
  \multicolumn{3}{c|}{\textbf{Na\"ive Bayes}} &
  \multicolumn{3}{c|}{\textbf{Logistic Regression}} \\ \hline
 &
   &
  \textbf{MR} &
  \textbf{ACC} &
  \textbf{MR} &
  \textbf{ACC} &
  \textbf{MR} &
  \textbf{ACC} &
  \textbf{\#train} &
  \textbf{MR} &
  \textbf{ACC} &
  \textbf{\#train} &
  \textbf{MR} &
  \textbf{ACC} \\ \hline
\multirow{6}{*}{Word2vec} &
  \multirow{2}{*}{50} &
  \multirow{2}{*}{2.97} &
  \multirow{2}{*}{0.18} &
  \multirow{2}{*}{2.67} &
  \multirow{2}{*}{0.27} &
  \multirow{2}{*}{2.54} &
  \multirow{2}{*}{0.27} &
  100 &
  2.09 &
  0.48 &
  100 &
  1.97 &
  0.48 \\ \cline{9-14} 
 &
   &
   &
   &
   &
   &
   &
   &
  166 &
  2.18 &
  0.48 &
  166 &
  1.97 &
  0.52 \\ \cline{2-14} 
 &
  \multirow{2}{*}{100} &
  \multirow{2}{*}{2.73} &
  \multirow{2}{*}{0.21} &
  \multirow{2}{*}{2.24} &
  \multirow{2}{*}{0.36} &
  \multirow{2}{*}{2.33} &
  \multirow{2}{*}{0.33} &
  100 &
  1.56 &
  0.64 &
  100 &
  1.84 &
  0.52 \\ \cline{9-14} 
 &
   &
   &
   &
   &
   &
   &
   &
  166 &
  1.51 &
  0.61 &
  166 &
  1.76 &
  0.54 \\ \cline{2-14} 
 &
  \multirow{2}{*}{300} &
  \multirow{2}{*}{2.58} &
  \multirow{2}{*}{0.33} &
  \multirow{2}{*}{2.48} &
  \multirow{2}{*}{0.24} &
  \multirow{2}{*}{2.27} &
  \multirow{2}{*}{0.30} &
  100 &
  1.70 &
  0.61 &
  100 &
  1.82 &
  0.52 \\ \cline{9-14} 
 &
   &
   &
   &
   &
   &
   &
   &
  166 &
  1.70 &
  0.64 &
  166 &
  1.76 &
  0.54 \\ \hline
\multirow{2}{*}{BERT} &
  \multirow{2}{*}{768} &
  \multirow{2}{*}{2.61} &
  \multirow{2}{*}{0.33} &
  \multirow{2}{*}{2.45} &
  \multirow{2}{*}{0.39} &
  \multirow{2}{*}{2.5} &
  \multirow{2}{*}{0.36} &
  100 &
  2.06 &
  0.52 &
  100 &
  1.97 &
  0.48 \\ \cline{9-14} 
 &
   &
   &
   &
   &
   &
   &
   &
  166 &
  2.12 &
  0.45 &
  166 &
  1.88 &
  0.54 \\ \hline
\end{tabular}
\caption{Performance on subset 2 (MR = mean rank, ACC = accuracy)}
\label{tab:subset2}
\end{table*}

\subsubsection{System 1} In this system, we use Word2vec word-embeddings of dimension 100 in the embedding layer. In the classification layer, we use L2 norm for subset 1 and Bernoulli Naive Bayes classifier for subset 2. This is the system that stood $1^{st}$ in the FinSim task in terms of both Mean Rank and Accuracy.
\subsubsection{System 2} In this system, we use Word2vec word-embeddings of dimension 300 in the embedding layer. In the classification layer, we use L2 norm for subset 1 and Bernoulli Naive Bayes classifier for subset 2.

\subsubsection{System 3} In this system, we use word-embeddings obtained from BERT of dimension 768 in the embedding layer. In the classification layer, we use logistic regression for both the subsets. 

\section{Results}
\label{sec:results}
We discuss the performance of all the approaches and systems on the test dataset.  Table \ref{tab:subset1} describes the results of different approaches on subset 1. It is clear from the table that unsupervised approaches (cosine similarity, L1 norm and L2 norm) prove to be better than supervised approaches (Na\"ive Bayes and Logistic Regression) for subset 1 with Word2vec word-embeddings. Among the unsupervised, L2 norm dominates. For BERT embeddings, logistic regression dominates. 

Table \ref{tab:subset2} describes the results of different approaches on subset 2. Contrary to subset 1, supervised approaches perform better than unsupervised approaches on subset 2. Na\"ive Bayes dominates among the supervised classifiers for the Word2vec word-embeddings while logistic regression dominates for BERT embeddings. Since we obtain 100\% accuracy for subset 1, as assumed based on the rule, and the training dataset is small, we add the terms in subset 1, with their predicted labels, in the training dataset. Hence, we present results for subset 2 on 100 training data points (original train dataset) as well as on 166 training data points (original train dataset + subset 1). In the following systems, we use results with 166 training data points on subset 2 for consistency.    

Table \ref{tab:finalres} shows the results for the systems discussed in subsection \ref{subsec:systems}. System 1 and 2 show the performance of Word2vec word-embeddings of dimensions 100 and 300, respectively. These systems are a combination of unsupervised and supervised approaches separately applied on subset 1 and 2, respectively. They outperform any of the approaches applied to the aggregate test data. For both the systems, the classification layers consist of L2  norm for subset 1 and Na\"ive Bayes classifier for subset 2 as they dominate in their respective categories. System 3, reveals the performance of BERT word-embeddings in the embedding layer. It uses the logistic regression classifier, for both subsets 1 and 2, in the classification layer as it performs the best with BERT word-embeddings.

\begin{table}[H]
\centering
\renewcommand{\arraystretch}{1.33}%
\begin{tabular}{|c|c|c|}
\hline
\textbf{System} & \textbf{Mean Rank} & \textbf{Accuracy} \\ \hline
1               & 1.17               & 0.87              \\ \hline
2               & 1.23               & 0.88              \\ \hline
3               & 1.29               & 0.85              \\ \hline
\end{tabular}
\caption{Results of different systems on the whole test data}
\label{tab:finalres}
\end{table}

Although system 1 stood $1{st}$ in the task on both metrics, in post-submission analysis, system 2 outperforms system 1 in terms of accuracy. Overall, the Word2vec embeddings outperform BERT embeddings. This may be because BERT embeddings are context-dependent and do not produce a unique embedding for each word. On the contrary, Word2vec embeddings are unique for every word and are more suited for a task where proper nouns are being classified. 


\section{Conclusion}
As part of FinSim 2020 shared task on Learning Semantic Representations in the Financial Domain, we attempt to solve the problem of hypernym detection minted for Financial texts. We employ static Word2vec and dynamic BERT embeddings under the top classification layers consisting of simple classifiers. Word2vec dominates for both dimensions (100 and 300). Though BERT embeddings come out to be equally accurate for terms containing the one hypernym within them, they lag behind for the other subset of terms. With higher computational resources, BERT could be pre-trained on the whole corpus, and the performance may improve. Unsupervised metrics are efficient and independent of data size, but they lag behind supervised classifiers for terms exclusive of class label. 

For future research, the data size could be increased significantly to bring deep learning based classifiers into the picture, and the task could be enhanced from hypernym detection to hypernym discovery. Overall, the task advances the NLP community towards the broad area of Financial Document Processing and encourages collaboration between the fields of Finance and NLP. 
\label{sec:conclusion}

\bibliographystyle{named}
\bibliography{ijcai20}
\end{document}